\documentclass[sigconf, nonacm, screen=true, review=false, table=true, anonymous=false, natbib=false]{acmart}
\AtBeginDocument{%
  }

\RequirePackage[
  datamodel=acmdatamodel,
  style=acmnumeric,
  sorting=none
  ]{biblatex}

\usepackage{pifont}
\newcommand{\xmark}{\ding{55}}

\usepackage{mathtools}
\usepackage{braket}
\DeclareMathOperator*{\argmin}{arg\,min}

\usepackage[noline,noend]{algorithm2e}
\SetNlSty{}{}{}
\let\oldnl\nl
\newcommand\nonl{%
  \renewcommand{\nl}{\let\nl\oldnl}}
\SetKwComment{Comment}{/* }{ */}

\begin{document}

\title{Distributed Equivariant Graph Neural Networks for Large-Scale Electronic Structure Prediction}

\author{Manasa Kaniselvan}
\email{mkaniselvan@ethz.ch}
\affiliation{
  \institution{ETH Zurich}
  \city{Zurich}
  \country{Switzerland}
}

\author{Alexander Maeder}
\affiliation{%
  \institution{ETH Zurich}
  \city{Zurich}
  \country{Switzerland}}
\email{almaeder@ethz.ch}

\author{Chen Hao Xia}
\affiliation{%
  \institution{ETH Zurich}
  \city{Zurich}
  \country{Switzerland}
}
\email{chexia@iis.ee.ethz.c}

\author{Alexandros Nikolaos Ziogas}
\affiliation{%
 \institution{ETH Zurich}
 \city{Zurich}
 \country{Switzerland}}
\email{alziogas@iis.ee.ethz.ch}

\author{Mathieu Luisier}
\affiliation{%
  \institution{ETH Zurich}
  \city{Zurich}
  \country{Switzerland}}
\email{mluisier@iis.ee.ethz.ch}

\begin{abstract}

Equivariant Graph Neural Networks (eGNNs) trained on density-functional theory (DFT) data can potentially perform electronic structure prediction at unprecedented scales, enabling investigation of the electronic properties of materials with extended defects, interfaces, or exhibiting disordered phases. However, as interactions between atomic orbitals typically extend over 10+ \AA, the graph representations required for this task tend to be densely connected, and the memory requirements to perform training and inference on these large structures can exceed the limits of modern GPUs. Here we present a distributed eGNN implementation which leverages direct GPU communication and introduce a partitioning strategy of the input graph to reduce the number of embedding exchanges between GPUs. Our implementation shows strong scaling up to 128 GPUs, and weak scaling up to 512 GPUs with 87\% parallel efficiency for structures with 3,000 to 190,000 atoms on the Alps supercomputer. 
\end{abstract}

  
\maketitle

\section{Introduction}

The electronic structure of a material, encoded in its ground-state Hamiltonian matrix, provides information about the spatial and energetic distribution of states that electrons can occupy in it. Although not a direct observable, the Hamiltonian matrix serves as an input to, for example, quantum transport (QT) simulations that aim to compute the electronic current flowing through devices when an external voltage bias is applied \cite{Brandbyge2002}. Advanced QT codes can now handle systems with 10k+ atoms while accounting for realistic many-body effects \cite{Deuschle2024}. As such, they are approaching the physical dimensions of modern semiconductor devices, in particular transistors, the active components of all integrated circuits (IC) \cite{Yeap2024}. The ability to operate on such scales allows QT solvers to serve as powerful tools to evaluate new device designs at atomic resolution and predict their performance in future IC technologies.

Currently, density-functional theory (DFT) is widely used to compute the electronic structures of almost all kinds of materials. This framework implements the Kohn-Sham equations, which self- consistently couple the electron density and electrostatic potential of the atomic system of interest until convergence is achieved, after $N_{iter}$ iterations \cite{Kohn1965}. Several commercially or freely available codes implement this self-consistent and iterative ``density$\leftrightarrow$potential'' scheme on input atomic structures, and are in themselves high-performance computing (HPC) applications. The core operation of DFT consists of repeatedly solving eigenvalue problems for matrices of size $N_{orb}\times N_{orb}$, where $N_{orb}$ is the total number of atomic orbitals in the system. The computational complexity of the full method scales with \(\mathcal{O}(N_{iter}N_{orb}^3)\), making it highly intensive in structures with thousands of atoms. 

Due to the computational cost of running DFT at scale, large-scale QT simulations are no longer limited by the time necessary to solve the underlying Schr\"odinger equation with open boundary conditions \cite{Ziogas2019}, but by the ability to generate the required input \textit{ab-initio} Hamiltonian matrices at the same scales \cite{Brandbyge2002}. The simplest device structures, such as homogeneous nanowires made of crystalline silicon, can be represented by tiling the Hamiltonian matrix with a small unit cell which is computationally manageable to generate. However in case of surface roughness, poly-crystallinity, non-uniform strain, atomic disorder, or randomly placed defects, this approach breaks down \cite{Liu2024}, and large Hamiltonians corresponding to the whole device geometry must be produced at the DFT level of accuracy. Depending on the considered sizes, doing so might be computationally out of reach, excluding many realistic systems from computational investigations of their transport properties.

\begin{figure}[!t]
  \centering
  \includegraphics[width=\linewidth]{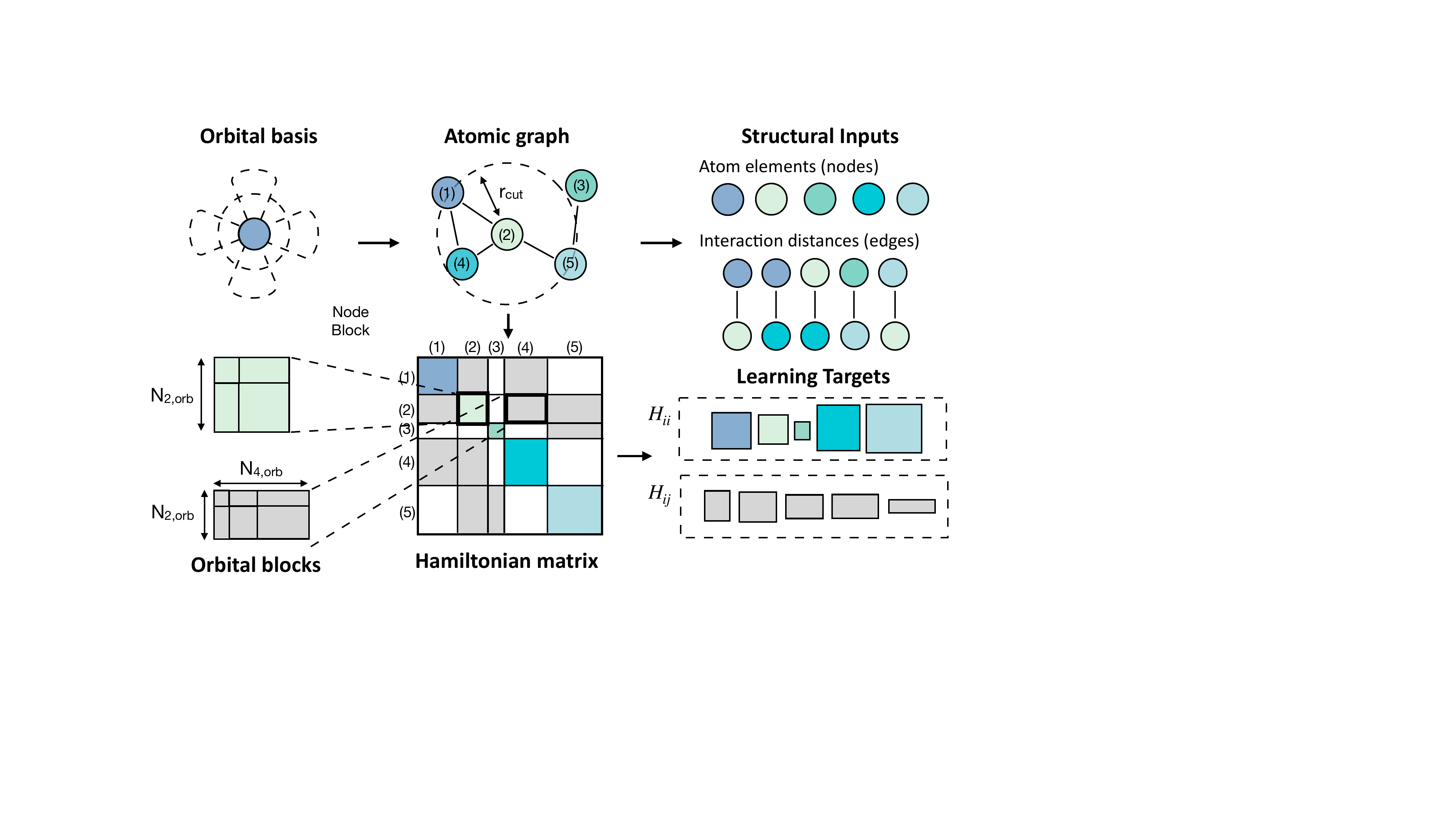}
  \caption{Overview of the electronic structure prediction problem, where the mapping between atoms/bonds and blocks of the Hamiltonian matrix is learned.}
  \label{fig:intro}
\end{figure}

To bridge the gap between the computational demands of DFT and the scale requirements of device-level simulations, one promising approach lies in training Graph Neural Networks (GNNs) to learn the mapping between the atomic and electronic structures of materials \cite{quantumham, Wang2024, deeph3}. This `electronic structure prediction' (ESP) problem involves learning the Hamiltonian matrix $\mathbf{H}$ as a function of the atomic positions. Given a set of atomic positions as input, the prediction targets in ESP are the sub-matrices $\smash{H_{ij}^{\alpha \beta}}$ of $\mathbf{H}$ expressed in a localized basis set such as Gaussian-type orbitals (GTO) \cite{cp2k, orca}, as illustrated in \textbf{Fig.~\ref{fig:intro}}. Each sub-matrix contains the interactions between orbitals $\alpha$ and $\beta$ on atoms $i$ and $j$, and is covariant to global rotation of the atoms. Almost all ML ESP models are designed to maintain this rotational equivariance within all internal network operations, such that global rotations of the input structure produce correctly rotated output matrices \cite{Batzner2022, eSEN}. Once these networks are trained, they can be used to efficiently construct the Hamiltonian matrix of large samples with thousands of atoms in \(\mathcal{O}(N_{orb})\) time, within a single inference step \cite{Behler2016}.

These rotationally equivariant Graph Neural Networks typically use spherical representations of the nodes and edges. Each feature thus requires an additional dimension $l$ to encode the dependence of each learnable channel on the angular degree. On top of that, in ESP networks, the connectivity of the graph, often defined in terms of a cutoff radius (r$_{cut}$), is not entirely a hyperparameter but rather constrained by the distances which encompass a sufficient fraction of nonzero orbital interactions in $\mathbf{H}$. In practice, r$_{cut}$ tends to be greater than 10 \AA\; in many relevant materials, leading to densely-connected graphs with hundreds of edges for every node. The combination of increased feature dimension and dense connectivity leads the memory consumption of these networks to be prohibitively high for large atomic structures. Most existing ML ESP models are thus limited to graph sizes of $<$ 150 atoms \cite{wanet, universal_materials}. Computational challenges arise when treating large, densely connected graphs where both nodes ($\smash{H_{ii}^{\alpha \beta}}$) and edges ($\smash{H_{ij}^{\alpha \beta}}$) must be learned.

To tackle this challenge, we (1) develop a distributed Graph Neural Network (GNN) application for electronic structure prediction, and (2) demonstrate an efficient partitioning approach to reduce communication overhead when treating structures represented by large graphs. Together, these computational optimizations enable training on structures with over an order of magnitude more atoms than previously treated, and permit the use of the large r$_{cut}$ required for ESP. As an example, we consider an amorphous (disordered) material made of 3,000 atoms (nodes) and 500,000$+$ inter-atomic orbital interactions (edges). When processing it our application scales up to 128 GH200 GPUs (32 nodes) on the Alps supercomputer at the Swiss National Supercomputing Centre (CSCS). Starting from the same structure, we achieve a weak scaling efficiency of 87\% on 512 GPUs (128 nodes). Our application can be combined with advanced quantum transport approaches to explore the electronic properties of devices with non-ideal structures, such as phase-change memory \cite{Zhou2023}, silicon nano-ribbon field-effect transistors \cite{Yeap2024}, or nano-ionic resistive memories \cite{Onofrio2015, Kaniselvan2023}. 

\section{Background \& Related Work}
\label{sec:background}

\subsection{Equivariant GNNs in materials science}

GNNs have become mainstream in computational materials science, where they are used to learn structure-property relationships on atomic graphs. Training in these GNNs occurs through message passing, where nodes and/or edges of the graph are updated as a learnable function of other nodes/edges (typically atomic elements/interatomic distance) in their neighborhoods \cite{Gilmer2017}. Successive message passing layers gradually transform the initial identity of each node/edge such that they can be mapped to targets representing material properties. 

The most commonly treated prediction targets are atomic forces. This is the basis behind Machine Learned Force Fields (MLFFs), which have received a wide attention from the machine learning (ML) and materials science communities over the last 10 years \cite{Behler2016}. After training, MLFFs are used re-compute forces along molecular dynamics (MD) trajectories, enabling simulations of systems with tens of millions of atoms \cite{deepmd}.

The success of GNNs for atomic forces has recently motivated their extension to electronically resolved properties, which are a function of the atomic orbitals available in the structure. The reasons are the following: First, the \textit{ab-initio} training data used for MLFFs comes from DFT, and is expensive to generate. ESP networks can be leveraged to generate efficient initial guesses to accelerate the production of this training data \cite{wanet}. Second, atomic forces are based on underlying electronic properties. Directly encoding those electronic properties is a promising approach to improve the accuracy when predicting forces \cite{Suman2025}. Finally, as previously motivated here, simulation methods such as quantum transport directly require the matrix-representation of electronic structure of the materials they treat \cite{Brandbyge2002}. 

This ESP task is, however, more complex than the prediction of atomic- and molecular-properties. Unlike force-prediction, which requires per-node outputs, networks tailored for electronic structure prediction must include learnable embeddings for all the edges of the graph, which correspond to interactions between orbitals on different atoms. This number of edges is typically orders of magnitude larger than the number of nodes. In addition, predicting orbital blocks requires higher-degree tensor representations in both node and edge features. This increases the dimensionality of the embeddings and significantly raises the memory consumption of the input data. Rotational equivariance in such models can be maintained by using tensor products to mix features \cite{TFNs}. Compared to otherwise rotationally `invariant' networks, which depend on data-augmentation to learn these relationships, architecturally group-equivariant networks are known to learn physically-meaningful mappings with less data \cite{Batzner2022} and lower compute budgets \cite{brehmer2024doesequivariancematterscale}. However, these tensor product operations scale poorly with \(\mathcal{O}(l_{max}^6)\), the key physical parameter $l_{max}$ being the maximum degree of the angular momentum of the spherical harmonic basis used as embeddings. High values of $l_{max}$ ($\geq$4) are required to adequately fit material regions where the electron wavefunctions vary sharply within small distances, such as in the case of interactions between $d$-orbitals. Early implementations of equivariant GNNs for ESP have thus been limited to small molecules (< 30 atoms \cite{quantumham}).

These tensor products can, however, be simplified to linear layers through local rotations of edges to align to a fixed axis \cite{so2}. Several ESP models have recently adopted this mathematical trick to reduce the complexity of the network with respect to l$_{max}$ to \(\mathcal{O}(l_{max}^3)\), and enable the treatment of materials with 150 to 1,000 atoms \cite{wanet, amorphous}. The possibility to treat such large atomic structures could unlock unprecedented insight into the electronic properties of materials and devices.

\subsection{The electronic structure problem}
\label{sec:esp}
The ground-state electronic structure of a material is determined by solving the Schr\"odinger equation, which is the central component of many DFT codes \cite{Kohn1965}. Several practical implementations operate with a basis of localized atomic orbitals $|\varphi\rangle$, often constructed from contracted Gaussian functions \cite{cp2k, orca}. The Schr\"odinger equation then takes the form of a generalized eigenvalue problem: \(\textbf{H} \psi = \varepsilon \textbf{S} \psi\). Here, the Hamiltonian matrix $\mathbf{H}$ has entries $H_{ij}=\langle \varphi_i | \hat{H}(\mathbf{r}) | \varphi_j \rangle$ where $\hat{H}(\mathbf{r})$ is the Hamiltonian operator. The overlap matrix $\mathbf{S}$ is made of terms $S_{ij}=\langle \varphi_i | \varphi_j \rangle $ that describe finite overlaps between the localized orbitals, and can be pre-computed from the orbital basis.
Both $\mathbf{H}$ and $\mathbf{S}$ are coarse-grained matrices of size \(N_{orb} = \sum_k N_{A}^k \cdot N_{A, orb}^k \), where \(N_{A}\) is the number of atoms, \(N_{A, orb}\) the number of orbitals per atom, and $k$ indexes over the different atomic species found in the system. The solution of the Schr\"odinger equation provides the energy eigenvalues ($\varepsilon$) and wavefunctions ($\psi$). Together, they describe states that electrons can occupy within a material. For a fixed basis, there exists a mapping  \( F: \{ r_i, Z_i \} \rightarrow \mathbf{H} \) between the coordinates ($\mathbf{\{r_i\}}$) and identity (atomic numbers $\{Z_i\}$) of the atoms in a material and its Hamiltonian matrix \cite{Hohenberg1964}. 

The entries of $\mathbf{H}$ are covariant to rotations of the input atomic structure. This property emerges from the rotational symmetries of the underlying localized orbitals in which $\psi$ are expanded. Specifically, they resemble the spherical harmonics \( Y^{l}_m(\hat{r}')\), each specified by an angular momentum degree $l$ and order \( m \in \{-l, \dots, l\} \). Under a specific rotation \( R \), each orbital transforms according to the Wigner-D matrix of the corresponding degree $l$ for that rotation: \( Y^{l}_m(\hat{r}') = \sum_{m'} D^{l}_{mm'}(R) Y^{l}_{m'}(\hat{r})\) where \( \hat{r} \)/\( \hat{r}' \) are the normalized direction vectors before/after the rotation. 

The subblocks of $\mathbf{H}$, $\smash{\mathbf{H}^{\alpha\beta}_{ij}}$, contain the interactions between the spherical harmonic basis functions of degree $l_\alpha$ on atom $i$ and degree $l_\beta$ on atom $j$. Mathematically, it can be described as the tensor product $l_\alpha \otimes l_\beta$ of length $(2l_\alpha + 1)\times (2l_\beta + 1)$ between the uncoupled angular momentum eigenstates $l_\alpha$ and $l_\beta$. Each of these tensor products can be decomposed into a direct sum ($\oplus$) of coupled angular momentum eigenstates $\ket{L, M}$, where $L$ ranges from $|l_\alpha-l_\beta|$ to $|l_\alpha+l_\beta|$: $T(l_\alpha \otimes l_\beta)  = |l_\alpha-l_\beta|\oplus...\oplus (l_\alpha+l_\beta)$. This transformation $T$ is performed via a matrix of Clebsch-Gordon coefficients. For example, the coefficient for a specific $\ket{L, M}$ coupled angular momentum eigenstate component is given by:

\begin{equation}
\ket{L, M} = \sum_{m_\alpha=-l_\alpha}^{+l_\alpha}\sum_{m_\beta=-l_\beta}^{+l_\beta}C_{(l_\alpha, m_\alpha)(l_\beta, m_\beta)}^{(L, M)}\ket{l_\alpha, m_\alpha}\ket{l_\beta, m_\beta}.
\end{equation}

\noindent Here, $C_{(l_\alpha, m_\alpha)(l_\beta, m_\beta)}^{(L, M)}$ is the Clebsch-Gordon coefficient describing the contribution of $\ket{l_\alpha, m_\alpha} \otimes \ket{l_\beta, m_\beta}$ to the coupled state $\ket{L, M}$. The transformation $T$ thus decomposes the tensor product $\smash{\mathbf{H}^{\alpha\beta}_{ij}}$ into a basis of coupled angular momentum states, where each subspace corresponds to a specific total angular momentum $L$. Applying a rotation $\mathbf{R}$ to the resulting $\smash{\mathbf{H}^{\gamma\delta}_{ij}}$ then consists of applying the corresponding Wigner-D transformation \(D^{l}_{mm'}(R)\) to each of its subspaces of degree $l$. 

\subsection{Related work}
\setlength{\tabcolsep}{3pt} 
\begin{table}[]
\begin{tabular}{@{}lllllll@{}}
\toprule
\textbf{Application}& \textbf{Sym.}& \textbf{Task}& \multicolumn{2}{c}{\textbf{Prediction}} & \textbf{Distr.} &  $\mathbf{N_A}$\\
 &  &  & \textbf{Node} & \textbf{Edge} &  &  \\
 \midrule
DeepMD \cite{deepmd}   & None      & F \& E      &   \checkmark&   \xmark&  \checkmark          &  -\\
Allegro \cite{Kozinsky2023}              & SO(3)                            & F \& E            &   \checkmark&   \xmark&  \checkmark               &  -\\
MACE \cite{mace}   & SO(3)      & F \& E      &   \checkmark&   \xmark&  \xmark          &  -\\
SevenNet \cite{Park2024}   & SO(3)     & F \& E      &   \checkmark&   \xmark&  \checkmark&  -\\
EquiformerV2 \cite{equiformerv2}   &  \textbf{SO(2)}& F&   \checkmark&   \xmark&  \xmark          &  -\\
QHNet \cite{quantumham}   & SO(3)      & $H_{ij}$&   \checkmark&   \checkmark&  \xmark          &  10\\
 WANet \cite{wanet}& \textbf{SO(2)}& $H_{ij}$& \checkmark& \checkmark& \xmark&100\\
DeepH2 \cite{deeph2, Wang2024}   & \textbf{SO(2)}& $H_{ij}$&   \checkmark&   \checkmark&  \xmark          &  150 \\
\midrule
\textbf{This work}                  &  \textbf{SO(2)}& $H_{ij}$&   \checkmark&   \checkmark&  \checkmark          &  8000+\\ \bottomrule
\end{tabular}
\caption{Functional similarities and differences between our approach and other GNN applications for materials modeling, in terms of symmetry constraints, prediction tasks, and whether the application can leverage a distributed compute environment. F\&E = Forces \& Energies, $H_{ij}$ = Hamiltonian. Sym = Symmetry group, Distr = Distributed, $\mathbf{N_A}$ = Number of atoms.}
\label{tab:tab}
\end{table} 

\begin{figure*}[h]
  \centering
  \includegraphics[width=0.95\textwidth]{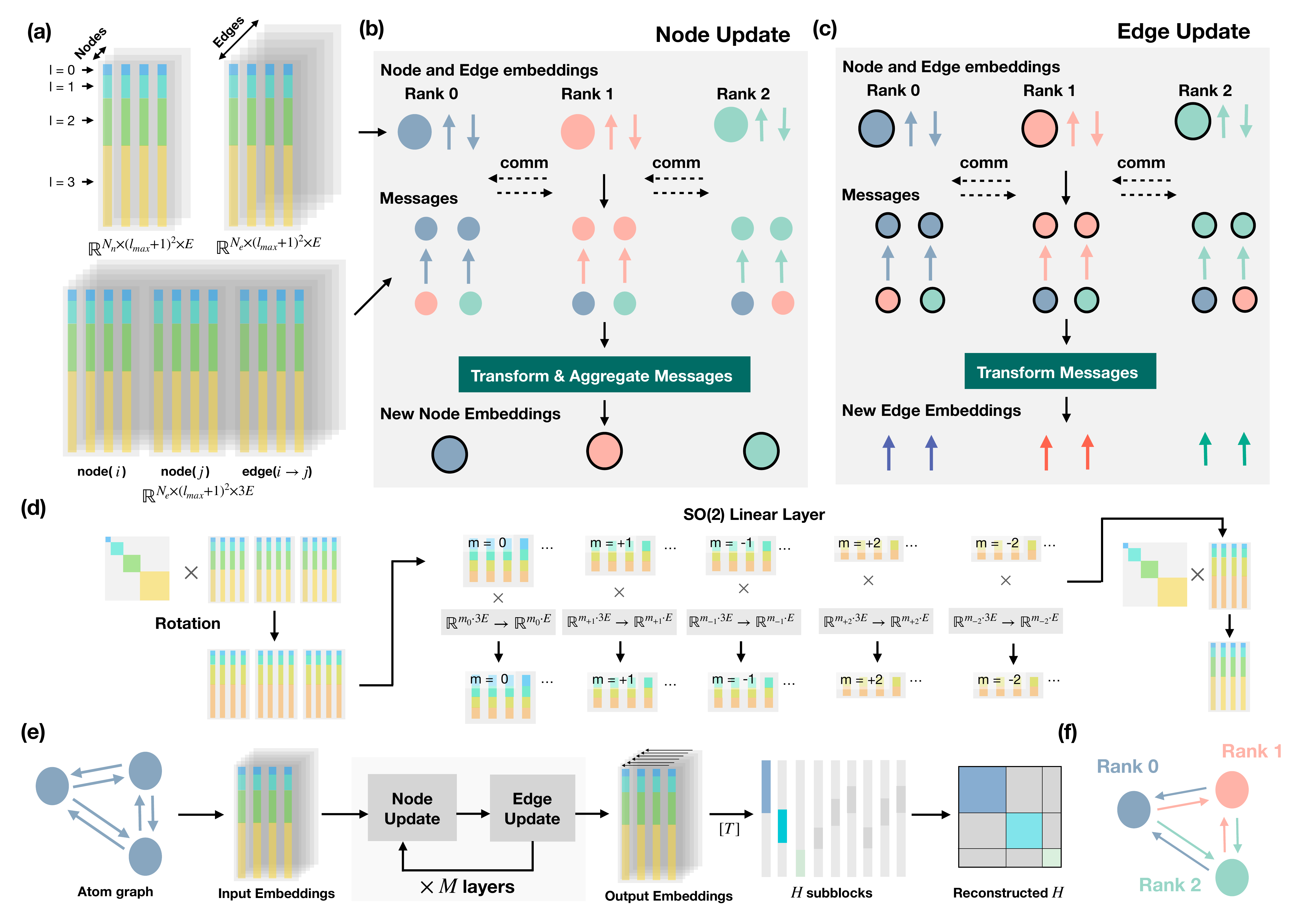}
  \caption{Overview of the network and data-distribution scheme. \textbf{(a)} (Top) Each node and edge of the graph is initialized with a learnable embedding composed of stacks of spherical tensors with length $l_{max}$, each expanded into an embedding dimension E. (Bottom) Each `Message' within the forward pass occurs between two nodes, and is the concatenation of the embeddings of the nodes and the edge connecting them. \textbf{(b)} Overview of the creation, transformation, and aggregation of messages to update the embeddings of each node. \textbf{(c)} The updated node embeddings are then used to update the embeddings of each edge. \textbf{(d)} Details behind the transformations which occur independently on every message: The embeddings are first rotated into a common axis through multiplication with the Wigner-D matrix ($W_D(\hat{r})$) defined for every bond edge. They are then decomposed into $m$-components which undergo separate linear layers, to mix features with different $l$ without breaking rotational equivariance. Finally, the transformed embeddings, with reduced embedding dimensionality $3E \rightarrow E$, are rotated back into their original bond axes with $W_D(\hat{r})^{-1}$. \textbf{(e)} This sequential update of node and edge embeddings occurs $M$ times, where $M$ is the number of message passing layers in the network, before the final embeddings are collapsed into the embedding dimension $E \rightarrow 1$ and transformed into the (padded) predicted outputs for each block of the Hamiltonian matrix. \textbf{(f)} Proposed `incoming edge' scheme to distribute a graph with 3 nodes/6 edges across three ranks (represented by the colors blue, red, yellow). Every rank owns one node, as well as both of its incoming edges. }
  \label{fig:network}
\end{figure*}

The similarities between our application and other GNNs for applications in computational materials science are summarized in \textbf{Table \ref{tab:tab}}. Here we also highlight the functional differences between MLFFs (node prediction task) and ML ESPs (node and edge prediction task). In the former, long-range information can be incorporated indirectly through message passing, while in the latter, the connectivity of the graph must explicitly include all relevant edges. For the networks performing similar electronic structure prediction, we additionally include the maximum system sizes that have been so far reported, and compare them to the atomic structure sizes enabled by our distributed application.

Most existing distributed GNN applications in materials science tackle the closely related problem of atomic-level force and energy prediction. Almost all these codes, e.g., DeepMD \cite{deepmd} and Allegro \cite{Kozinsky2023}, leverage the domain decomposition functionality of the LAMMPS molecular dynamics code \cite{LAMMPS}. LAMMPS relies on spatial decomposition of the domain into boxes, and has an additional functionality to modify the box-boundaries and balance the number of atoms within each box. During message passing iterations, ghost atoms are communicated through halo node/edges at the boundaries to aggregate messages onto each node. The communicated volume is determined by the connectivity of the graph - densely connected graphs incur high communication overhead as ghost atoms must send messages across multiple layers. 

Directly partitioning the graph rather than partitioning the spatial domain is a promising approach to improve load balancing and more evenly distribute communication. This is the approach taken by SevenNet \cite{Park2024, Batzner2022}, which is similarly integrated into LAMMPS. However, the distribution is generally performed only over nodes (atoms) and not over edges, making this approach less efficient for ESP where there are typically far more graph edges than nodes. Other approaches have been tailored for GNNs architectures with explicit higher-order interaction terms \cite{anuroop, distmlip}. On the other hand, networks like Allegro \cite{Kozinsky2023}, are designed to be strictly `local' - it considers interactions only within a fixed radius, such as 6 \AA\; \cite{Musaelian2023}. Higher parallelism is enabled by significantly reducing the amount of communication required. Nevertheless, it remains to be seen if such strictly local models generalize well to situations involving non-local interactions.

The acceleration of the specific class of equivariant GNNs has in large part focused on optimizing the expensive tensor product layers often used to enforce SO(3) rotational equivariance. Note that approaches using SO(2) layers \cite{wanet, equiformerv2, deeph2, amorphous}, as we use here, bypass the explicit computation of tensor products and thus do not encounter this specific computational challenge.


\section{Methods}
\label{sec:methods}

Our GNN application is based on the architectures of \textbf{Refs.}~\cite{amorphous, equiformerv2}. Here we describe the graph initialization and message passing operations. 


\textbf{Initialization of node and edge embeddings: } The graph $\mathcal{G}$ with nodes $\mathcal{N}$ and edges $\mathcal{E}$ is constructed with nodes corresponding to each atom $i$, and edges $i \rightarrow j$ within a specified r$_{cut}$. Each node $n_i$ is represented by a tensor $\smash{T^{node}_{lmfi}}$ with $l \in\{0, ..., l_{max}\}$, $m \in \{-l,...,l\}$, and  $f \in \{0,...,E\}$, where E is the size in which each $\ket{L, M}$ is embedded. The $lm$ dimensions are flattened in a single dimension $h$ of size $(l_{max}+1)^2$: $\smash{T^{node}_{lmfi} \Rightarrow T^{node}_{hfi}}$. The resulting structure of these embeddings are visualized in \textbf{Fig.~\ref{fig:network}(a)}. Hence, the total feature tensor for nodes has a size of $(l_{max}+1)^2 \times $E$ \times |\mathcal{N}|$. A similar feature tensor $\smash{T^{edge}_{hfk}}$ is created for edges with shape $(l_{max}+1)^2 \times $E$ \times |\mathcal{E}|$. Typically, $|\mathcal{E}|$ $\gg$ $|\mathcal{N}|$. At the start of training, the $L=0, M=0$ plane of $\smash{T^{node}_{hei}}$ is initialized with the atomic numbers $Z_i$ for each atom $i$, and the $L=0, M=0$ plane of $\smash{T^{edge}_{hfk}}$ are initialized with the edge distances expanded in a Gaussian basis. Rotation matrices along each bond edge (used for the SO(2) linear layers within the network) are also precomputed during this initialization step. 

\textbf{Message passing: } Within each message passing layer of the forward pass, the embeddings of nodes and edges are sequentially updated in two separate blocks (\textbf{Fig.~\ref{fig:network}(b)} and \textbf{Fig.~\ref{fig:network}(c)}). Both blocks begin by assembling a list of `messages' [M$_{i \rightarrow j}$], which are defined for every directed edge of index $k$ = $i \rightarrow j$ connecting node indices $i$ and $j$. Each message is the concatenation of the embeddings of the nodes $\smash{T^{node}_{hfi}}$, $\smash{T^{node}_{hfj}}$ and that of the edge $\smash{T^{edge}_{hfk}}$ between them, therefore having a shape of $\smash{M_{ghfk}}$ ($g \in \{0,1,2\}$). 

The message creation step therefore involves assembling lists of source node embeddings $\smash{T^{s}_{hfk}}$, target node embeddings $\smash{T^{t}_{hfk}}$, and edge embeddings $\smash{T^{e}_{hfk}}$ before finally concatenating them together in the $g$ dimension. The list of source/target node embeddings can be alternatively represented by a sparse tensor ($\smash{T^{s/t}_{hfk} \Rightarrow T^{s/t}_{hfij}}$) where the index $k$ is expanded to $ij$. This tensor in the $ij$ dimensions has the same sparsity pattern as the graph's adjacency matrix $\smash{A_{ij}}$. The operation to collect the list of source node embeddings is then equivalent to the sparse matrix multiplication $\smash{T^s_{hfij}}$ = $\smash{A_{ij} T^{node}_{hfjj}}$, where $\smash{T^{node}_{hfjj}}$ is a diagonal matrix containing the elements $\smash{T^{node}_{hfj}}$. Similarly, the target embeddings can be calculated from $\smash{T^t_{hfij}}$ = $\smash{T^{node}_{hfii} A_{ij}}$. Note that we use the term \textit{embedding} to refer to the tensors representing single nodes and edges of the atom-graph, and \textit{message} to refer to the concatenated [M$_{i \rightarrow j}$] that are transformed within the network. 

After the messages have been created, the forward pass through the network applies a pipeline of data transformations, rotations, and SO(2) linear layers to encode messages corresponding to each edge with information about their local atomic environments. The latter essentially applies a different linear layer to each order $m$ across embeddings of different $l$, implemented using a layout transformation of the data from $l$- to $m$-major ordering \cite{so2}. A simplified overview of the heavier operations performed on the messages is provided in \textbf{Fig.~\ref{fig:network}(d)}. The node update block ends with an aggregation step, consisting of a weighted sum over the incoming edges to every node. The learnable `attention' weights in this process encode the relative importance of each edge. In the edge update block, this step is omitted. Note that the number of messages corresponds to the number of edges, which thus defines the amount of data internally processed in both the node and edge update blocks during every forward pass. Further details behind the network architecture can be found in \textbf{Refs.} \cite{equiformerv2, amorphous}. 

\textbf{Figure \ref{fig:network}(e)} shows how the node and edge update blocks are connected within the model, where node/edge embeddings are sequentially updated through $M$ layers of message-passing to map the input atomic graph components to predicted (flattened) blocks of the Hamiltonian matrix $\mathbf{H}$. For training, $\mathbf{H}$ of the target structure is decomposed ahead of time into orbital blocks, and transformed into a coupled basis to form learning targets (\textbf{Section \ref{sec:esp}}). The graph's nodes are mapped to the diagonal Hamiltonian blocks $\smash{\mathbf{H_{ii}^{\gamma \delta}}}$ and the edges to the off-diagonal ones $\smash{\mathbf{H_{ij}^{\gamma \delta}}}$ ($i \neq j$). In the inference step, these predicted outputs can be post-processed back into the uncoupled basis with the reverse transformation: $T^{-1}(\smash{\mathbf{H}^{\gamma\delta}_{ij})}$ = $\smash{\mathbf{H}^{\alpha\beta}_{ij}}$ and used to reconstruct the Hamiltonian matrix.

\textbf{Distribution:} The way that nodes and edges are initially distributed between compute ranks determines the communication patterns required. If they are distributed naively, i.e., sequentially or randomly, between ranks, intermediate embeddings must be communicated during the message creation, attention mechanism, and aggregation steps. However, communication is unnecessary in the two last steps if each rank also owns the embeddings of all edges incoming to its nodes. This is illustrated in \textbf{Figure \ref{fig:network}(f)} with the simple example of a fully-connected 3-atom graph. When each rank owns the edges $i \rightarrow j$ which make up its set of messages [$i$, $j$, $i \rightarrow j$], it only needs to receive the incoming node embeddings $i$ from other ranks (as indicated with the dashed arrows labeled `comm' in \textbf{Figure \ref{fig:network}(b)-(c)}). Each rank can then independently perform the attention mechanism over its own edges to weigh incoming messages to its nodes ($j$). As communication becomes a bottleneck when the graph is finely partitioned, we adopt this `incoming edge' distribution strategy.


\section{Implementation}
\label{sec:implementation}

The near-sighted nature of electronic properties \cite{Kohn1996} allows us to apply a fixed cutoff radius r$_{cut}$ to define our graphs derived from input atomic structures. However, orbital interactions in electronic structure matrices can extend over 10+ \AA, as opposed to MLFFs where a typical value of r$_{cut}$ is 6 \AA\; \cite{Musaelian2023}. Hence, r$_{cut}$ must be chosen large enough to learn all relevant blocks $\mathbf{H_{ij}^{\alpha \beta}}$ between atoms $i$ and $j$. As a result, the number of messages handled during the forward pass is usually 2-3 orders of magnitude higher than the number of atoms. In contrast to conventional GNNs, the use of spherical harmonic representations also means that the tensor embedding of each node/edge has two dimensions, with size (l$_{max}$+1)$^2 \times$ E. Structures with thousands of atoms involve handling and communicating large volumes of data: for the structure of 3,000 atoms considered in \textbf{Section \ref{sec:optimization}}, the peak memory allocated within the forward pass during training is $\sim$115 GB. Distribution is thus necessary to be able to treat large graphs, leaving us with the challenge of handling dense communication patterns within our application.


\subsection{Hardware \& Software}

Our application was tested on the Alps supercomputer at CSCS, where every node has 4 NVIDIA GH200 Grace Hopper superchips fully connected with NVLink \cite{hoefler_alps_bench}. Every superchip combines a Hopper GPU with 132 Streaming Multiprocessors and 96 GB HBM3 memory with a Grace CPU with 72 cores and 128 GB LPDDR. The point-to-point bidirectional bandwidth between each Hopper GPU is 150 GB/s, resulting in 900 GB/s total bandwidth.

Our code is written in PyTorch \cite{pytorch2024} and CuPy \cite{cupy}, using torch tensors to represent the embeddings of nodes and edges. During training, we use torch's internal computational graph to construct gradients for the backward pass. We also use torch's Automatic Mixed Precision (torch.amp), which relies on float16 as the datatype during all numerically stable operations, taking advantage of the tensor cores of the GH200 hardware. This results in the embeddings (and thus the send/recv buffers for communication) being typically represented in the float16 format. The aggregation step is then performed in float32. To communicate embeddings between GPUs, we use CuPy's distributed functionality with the NVIDIA Collective Communications Library (NCCL) (cupyx.distributed $\rightarrow$ NCCLBackend) in combination with CuPy send/recv buffers. 
Each rank computes the loss for its local partition of the graph, followed by an `allgather' operation to collect the total loss across ranks. The backward pass is then performed on each rank, and communication of the gradients is handled by torch. Although here we focus on distribution of the input graph during the forward pass in order to treat larger structures during both training and inference, we note that methods such as the one proposed in \textbf{Ref.}~\cite{SAR} have been developed to partly forgo the communication of gradients in the backwards pass during training.


\begin{figure}[t]
  \centering
  \includegraphics[width=0.7\linewidth]{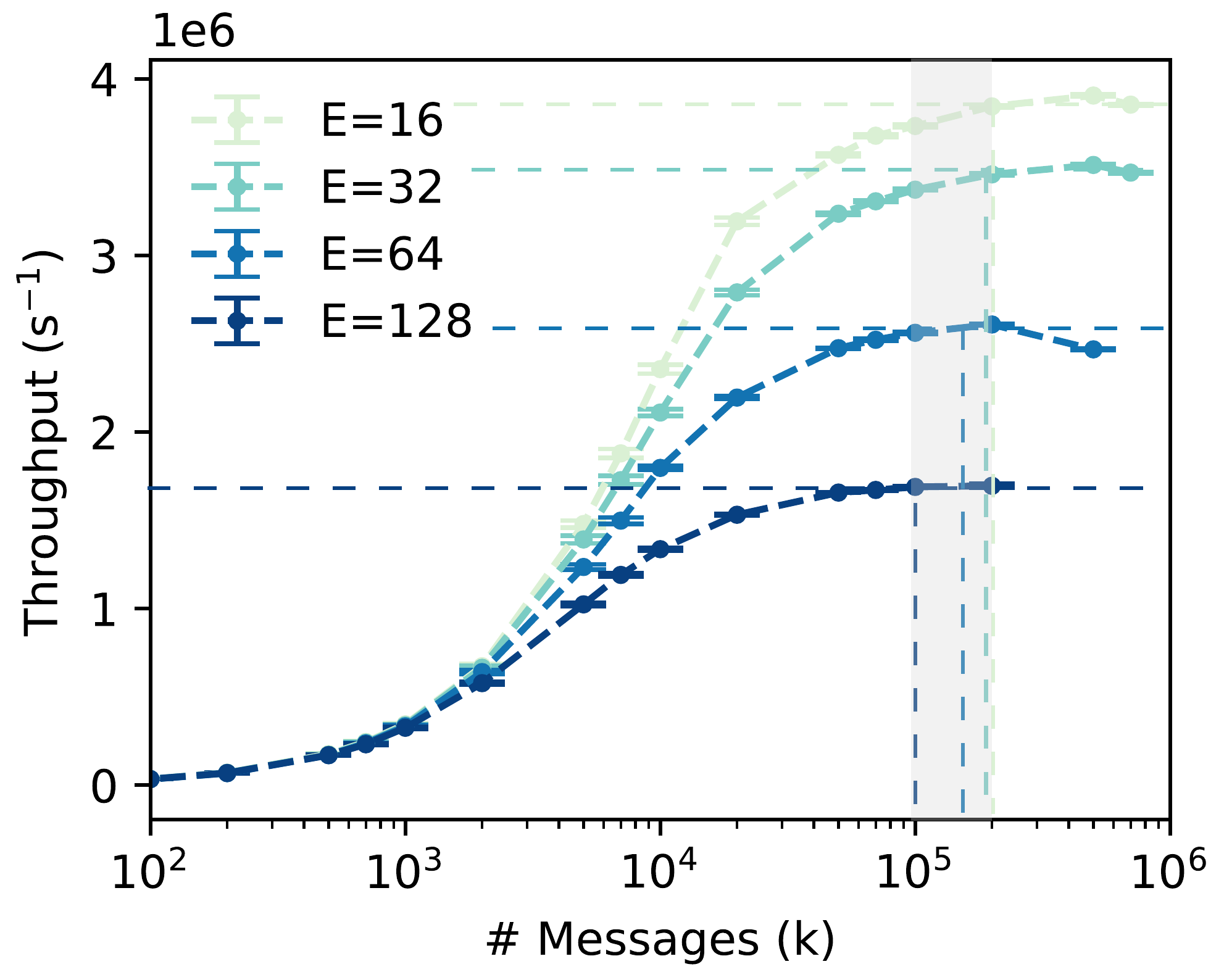}
  \caption{Message processing throughput for different embedding size E. The vertical dashed lines indicate the number of messages $k$ at which throughput is saturated for each E. Each data point corresponds to the median of 120 measurements (with 20 warm-up measurements).}
  \label{fig:batch_throughput}
\end{figure}

\begin{figure}[h]
  \centering
  \includegraphics[width=0.8\linewidth]{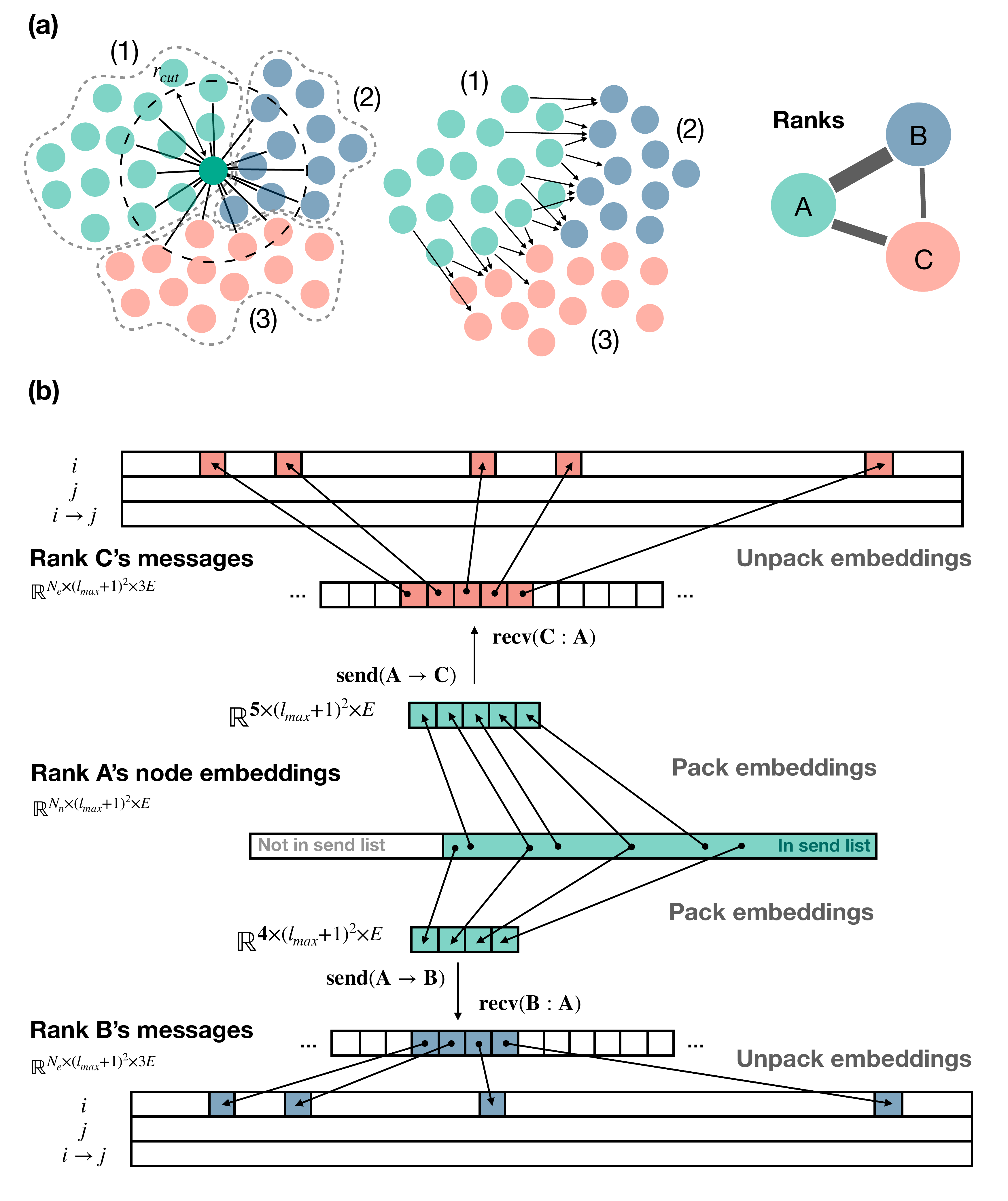}
  \caption{Communication patterns during the message creation. (a) (left) Schematic of a densely connected graph divided into three partitions, which are handled by three compute ranks. Edges are shown for one of the atoms in Partition (1). (center) Send operation during the message creation step, picturing the edges which Partition (1) sends to its neighboring partitions. (right) Schematic of the graph of compute ranks treating this problem, assuming that each rank owns one partition. The edge line widths indicate the number of embeddings which must be sent from one rank to another. (b) Schematic of the send/recv operation implementation. The node embeddings of each rank are ordered such that the ones to be sent to other ranks are grouped together. Embeddings from Rank A's (Partition 1) sub-list are randomly-accessed and packed into send buffers before being sent, resulting in only one send operation between any two ranks. In this setup, four (five) of the nodes required to assemble the messages of rank B (rank C) must be received from rank A.}
  \label{fig:memory_layout_comm}
\end{figure}

\subsection{Sustaining throughput}

The duration of the node/edge update blocks within the forward pass includes the times for message creation, transformation, and aggregation. The latter two steps are performed independently on the messages owned by each rank. The efficiency of these operations, in terms of throughput (messages processed per second) is highest when the data is large enough to saturate GPU utilization. To ensure that we remain within this regime when running on multiple ranks, the number of messages processed per rank should not drop below a specific amount. 

To quantify this, in \textbf{Fig.~\ref{fig:batch_throughput}}(a) we compute the achievable throughput of the message transformations as a function of the initial number of embeddings (i.e., the batch size) of size E. The messages processed have an input size of 3$\times$(l$_{max}$+1)$^2\times$E, and are initially stored in the float32 format. With E = 16 (E = 128), each message is 4.8 KiB (38.4 KiB). Internally, the message transformations involve a mix of matrix-vector products, element-wise operations, and projections to internally defined dimensions. The achievable throughput for this step begins to drop below 10$^5$ messages for E = 16. It is therefore not worth scaling beyond N$_{ranks}$ = $k$/2$\times$10$^5$ ranks with this value of E, since any speedup achieved by distribution would be counterbalanced by a loss in parallel efficiency during the remainder of the forward pass. We will use this finding as a guideline for our strong-scaling tests later on. 
 
\subsection{Communication during message creation}

As discussed in \textbf{Section \ref{sec:methods}}, under an `incoming edge' distribution scheme, only the initial construction of messages requires communication of node/edge embeddings between ranks. Taking the illustrated graph in \textbf{Fig. \ref{fig:memory_layout_comm}(a)}, an example of how this process is implemented between two given ranks during the message creation step is pictured in \textbf{Fig. \ref{fig:memory_layout_comm}(b)}. Note that the situation illustrated corresponds to a small r$_{cut}$ for visual clarity. In real systems, each atom can be connected to hundreds of neighbors, depending on the atomic configuration, and connections often extend beyond the 2-hop neighborhood when the graph is finely divided. The indices of atom-embeddings to be communicated are grouped at the end of the local atom-embedding lists of each rank. During the message creation step, all atom-embeddings belonging to a given rank $A$, which must be sent to another rank $B$, are packed into a send buffer. The nodes to be packed into each send buffer cannot be placed contiguously in memory, since the same node embedding must often be communicated to multiple ranks. The receive buffer on rank $B$ can be initialized ahead of time for a specific training structure.

\begin{figure*}[h]
  \centering
  \includegraphics[width=\textwidth]{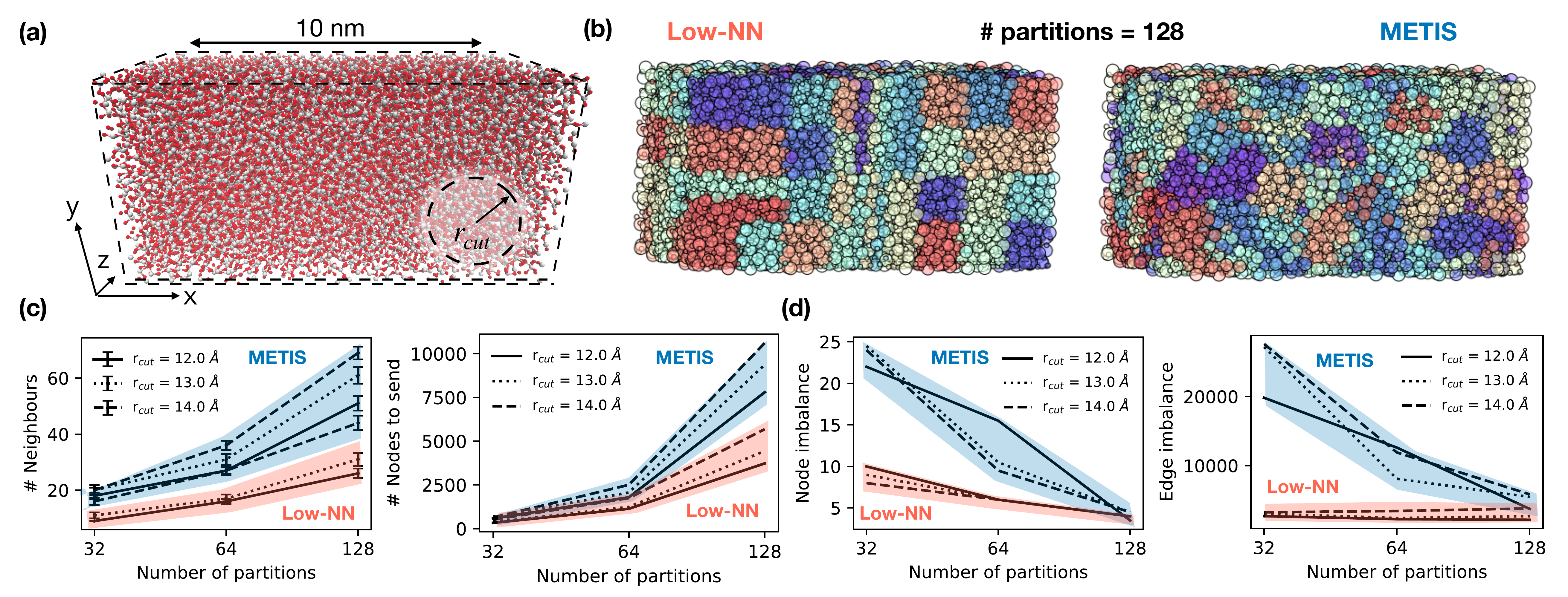}
  \caption{Partitioning input graphs. (a) Atomic structure of the HfO$_2$ unit cell considered in this work. It contains 24k atoms and 14.5 million edges. (b) Comparisons of the domains produced by the METIS (minimum-cut) and Low-NN (reduced-neighbor) algorithms when dividing the HfO$_2$ atomic graph with r$_{cut}$ = 12.0 \AA\; into 128 partitions (c) Anticipated communication overhead and (d) load imbalance arising from the applied partitioning methods when considering different r$_{cut}$. Node/edge imbalance is computed as $\max(\text{local nodes or edges}) / \text{mean}(\text{local nodes or edges})$, where `local’ refers to the number of nodes or edges assigned to each partition.}
  \label{fig:partitioning}
\end{figure*}

\begin{figure*}[t]
  \centering
  \includegraphics[width=\textwidth]{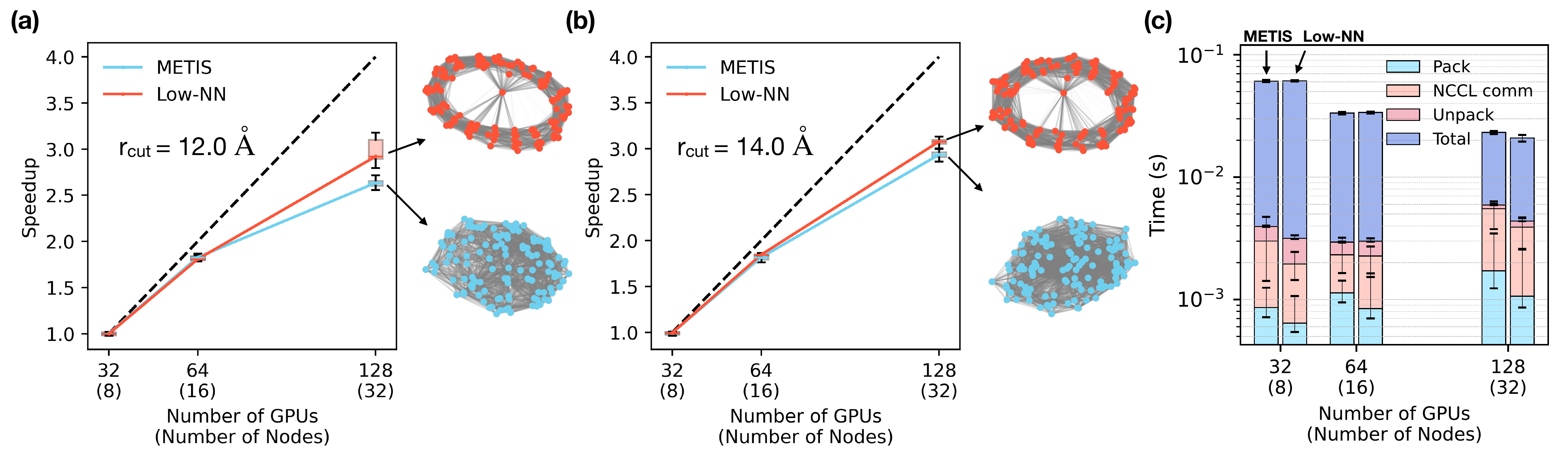}
  \caption{Strong scaling performance of the node update step under different graph distribution schemes, as a function of the cutoff radius r$_{cut}$, equal to (a) 12.0 \AA, and (b) 14.0 \AA, with embedding size E=16. The solid lines connect the boxplot medians of each data series. Scaling is plotted with respect to the best-performing distribution at 32 ranks. The dashed line represents ideal scaling, assuming that the message processing throughput remains saturated. To the side of each plot, we visualize the communication topology resulting from the METIS and Low-NN partitioning schemes at 128 GPUs. Each colored node corresponds to one compute rank. Edges indicate where communication takes place, and their thickness is proportional to the communication volume across the connection. (c) Time taken for communication overhead at each level of distribution, decomposed into the time taken to pack, send/recv, and unpack embeddings into the message tensors, following the layout of \textbf{Fig. \ref{fig:memory_layout_comm}(b)}. In (a)-(b)/(c), we measure runtimes separately for each rank, and show the median values over 120 epochs (with the first 20 discarded). The boxes/errorbars indicate the range (minimum to maximum) of runtime for different ranks.}
  \label{fig:reordering}
\end{figure*}

\subsection{Partitioning the input graph}
\label{sec:optimization}


With the constraint that every edge belongs to the same partition as its destination node, we now aim to partition the set of nodes and edges to both optimize load balancing and minimize the time spent on communication overhead. The latter includes (1) the time spent indexing and packing data into the send buffers (their size is equal to the number of neighboring ranks which each rank communicates with), (2) the time for the send and receive operations with NCCL, and (3) the time to unpack the received node embeddings and insert them into the message tensors. Steps (1) and (3) both involve random indexing operations to collect the required embeddings, as pictured in \textbf{Fig. \ref{fig:memory_layout_comm}(b)}. 

Communication overhead increases with the number of neighbors (NN) that each partition must send/receive node embeddings from, e.g., those within r$_{cut}$ of the boundary, because more buffers must be packed with random indexing operations, and send/recv postings show more latency. Rather than minimizing cuts between partitions, we are left with the modified problem of minimizing the number of neighbors created for each rank, while maintaining balanced partitions for the message-wise operations in the remainder of the forward pass. In some graphs, these objectives can be well-aligned, but in others they can be mutually exclusive, such as those with inhomogeneous density or a large distribution of node degrees. Graph representations of materials are also periodic, and edges extend across periodic boundaries. 

To address this, we develop a `Low-NN' algorithm which recursively bisects a graph between 2$^n$ ranks while (1) maintaining edge-wise balanced partitions, (2) accounting for 3-D periodicity, and (3) minimizing the number of new neighbors introduced by every local cut. Decisions on how to partition the sub-domain at each level are made to balance the total node degree (N$_D$) between the left and right partitions, as a proxy for balancing the number of edges. \textbf{Figure \ref{fig:partitioning}} illustrates the effect of our `Low-NN' partitioning approach. We start with the atomic graph pictured in \textbf{Fig. \ref{fig:partitioning}(a)}, which consists of an amorphous HfO$_2$ periodic unit cell of 3,000 atoms. The Hamiltonian $\mathbf{H}$ for this structure was computed with CP2K using a single-$\zeta$ valence (SZV) basis, which contains 10 orbitals per Hf atom and 4 orbitals per O atom. The size of the matrix is 18k $\times$ 18k. We then repeat the atomic structure $2\times$ along the $x$, $y$, and $z$ directions, such that its tiled graph representation contains 24,000 nodes and 14.5/22.8 million edges when connected to an r$_{cut}$ of 12.0 $\AA$/14.0 \AA. Running the Low-NN recursive bisection algorithm to a depth of 5, 6, and 7 generates 32, 64, and 128 partitions of the graph. In \textbf{Fig. \ref{fig:partitioning}(b)-(c)} we visualize the partitions generated at a depth of 7, compared to the same number of partitions found by the METIS algorithm \cite{metis}, which performs minimum-cut graph partitioning. The Low-NN partition results in a reduced number of neighbors for every rank, and fewer node embeddings that would need to be packed and communicated across partition boundaries, thus decreasing the expected communication overhead (\textbf{Fig. \ref{fig:partitioning}(c)}). It also tends to find balanced partitions (\textbf{Fig. \ref{fig:partitioning}(d)}). Partitioning overhead is easily amortized by training epochs, but we note that this approach is lightweight compared to METIS (8.54 sec for metis, 1.41 for low-NN to divide the structure in \textbf{Fig. \ref{fig:partitioning}(b)} into 64 partitions).

\begin{figure*}[t]
  \centering
  \includegraphics[width=\textwidth]{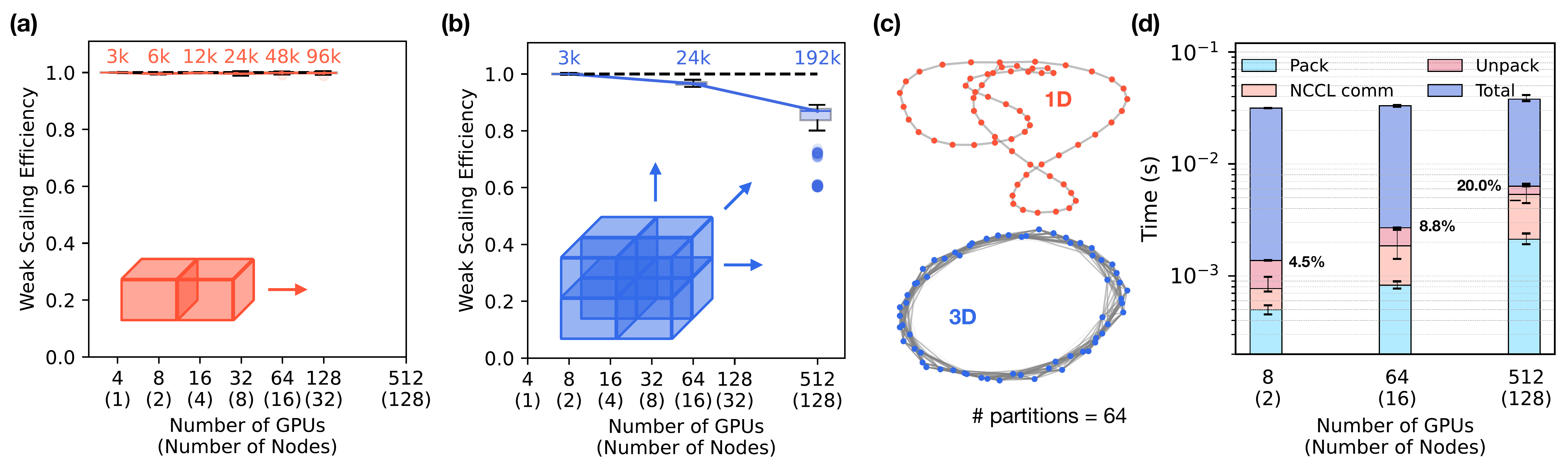}
  \caption{Weak scaling efficiency of the forward pass, considering a (a) 1-D (a) or (b) 3-D increase in problem size, as illustrated in the schematics within each plot. Each weak scaling experiment begins with the 3K atom HfO$_2$ structure introduced in Fig. \ref{fig:partitioning}(a). We then tile it in the specified dimension. The resulting number of atoms in the tiled structures are indicated above each data point. (c) Comparison of the communication topology resulting from 1-D (top) and 3-D (bottom) increases in problem size on 64 GPUs. The data behind the 3D weak scaling in (b) is additionally decomposed into (d) the time taken for communication overhead at each level of distribution, decomposed into the time taken to pack, send/recv, and unpack embeddings into the message tensors, following the layout of \textbf{Fig. \ref{fig:memory_layout_comm}(b)}. The percentages beside each bar indicate the fraction of runtime spent on this combined communication overhead. Similarly to Fig. \ref{fig:reordering}, error bars correspond to variation across ranks in the median forward pass time over forward passes of 120 epochs (discarding the first 20).}
  \label{fig:weakscaling}
\end{figure*}

We test the effect of partitioning on strong scaling efficiency in \textbf{Fig. \ref{fig:reordering}}, using the HfO$_2$ graph from \textbf{Fig. \ref{fig:partitioning}(a)}. Each partition is mapped to one GPU on Alps. As there are four GPUs per node, we indicate both the number of GPUs and nodes on the x-axis of each plot. Runtimes are measured for the node update block (\textbf{Fig. \ref{fig:network}(c)}), for two different values of r$_{cut}$ = 6.0 to 7.0 \AA , encompassing 99.3\% to 100\% of the nonzero elements of the corresponding $\mathbf{H}$. To compare with METIS, we use the implementation available in PyMETIS \cite{pymetis}. The communication topology resulting from each of these partitioning schemes at 128 ranks is visualized beside each plot. These topological plots were generated with the `spring layout' of the Python package networkx, which uses the Fruchterman-Reingold force-directed algorithm to position the nodes based on spring constants defined by the edge thicknesses \cite{Fruchterman1991}. In \textbf{Fig. \ref{fig:reordering}(c)} we decompose the total measured runtime in terms of the individual components contributing to communication overhead, following the process pictured in \textbf{Fig. \ref{fig:memory_layout_comm}}. Also indicated is the total measured runtime, the remainder of which accounts for compute-heavy operations such as the message transformation (\textbf{Fig. \ref{fig:network}(e)}) and aggregation steps.

When distributing the graph across $\leq$64 GPUs, the differences between the partitioning schemes are negligible as the total fraction of time spent on communication is low. At 128 GPUs, scalability becomes sensitive to the partition, and use of the low-NN algorithm improves the median achieved speedup by 10.8\% for r$_{cut}$ = 6.0\AA. Further scaling is limited by message throughput - each of the 128 ranks processing this graph already handles 1.1$\times$10$^5$ messages, and halving this would under-utilize the GPU (\textbf{Fig. \ref{fig:batch_throughput}}).
 
This trend can be predicted from the communication topology plots in \textbf{Fig. \ref{fig:reordering}(a)-(b)}. While the METIS algorithm minimizes cuts, it does not take into account the number of neighbors per node or the distribution of edges. It thus creates densely connected communication patterns that show high latency. Low-NN attempts to reduce the number of neighbors for each rank while also performing a finer, edge-wise distribution. With few ranks, it typically finds a 1-D-like split that optimally leverages the existence of periodic boundaries and avoids additional neighbors created by the existence of `corners' in the domain. However, at larger numbers of ranks, when it becomes favorable to introduce neighbors along other axes, it switches to a 2-D-like and eventually 3-D-like split. It can thus find an ideal ring-like communication pattern at all levels of distribution. This effect is also seen in the communication overhead plotted in \textbf{Fig. \ref{fig:reordering}(c)}. Due to the fewer neighbors per rank, the time taken to pack and send buffers with node embeddings is reduced. The NCCL communication time also shows a moderate improvement in the Low-NN case. At 128 ranks, when communication overhead represents a larger fraction of the total time, both effects together result in a reduction in runtime.

\section{Weak scaling efficiency on Alps}

Our goal is to process large graphs during both training and inference in order to learn the electronic structure of materials in the presence of structural or compositional disorder. While inference on large materials is straightforwardly motivated by the application, the necessity to train on them comes from the fact that disordered systems contain effects such as localized electronic states and long-range interactions which small unit cells cannot capture \cite{killians_paper, Liu2024, amorphous}. Learning accurate representation of such materials requires training data that adequately samples these disordered effects.

In practice, the size of the graph can increase in any dimension. Starting from the structure pictured in \textbf{Fig. \ref{fig:partitioning}(a)}, we therefore measure the weak scaling efficiency of the forward pass when the problem size (number of atoms) is increased by tiling in 1-D (\textbf{Fig. \ref{fig:weakscaling}(a)}, using the longest dimension $x$) or 3-D (\textbf{Fig. \ref{fig:weakscaling}(a)}). The graph is partitioned by the Low-NN algorithm introduced in \textbf{Section \ref{sec:optimization}}. Every data point is annotated with the total number of atoms in the corresponding tiled structure. We note that the material graph can be pre-initialized, and this is not included in the measured runtime. 

A 1-D increase in problem size maintains equal communication volume per rank, thus leading to the nearly perfect weak scaling efficiency of \textbf{Fig. \ref{fig:weakscaling}(a)}. However, this kind of weak scaling experiment does not mirror practical requirements, as the remaining two dimensions may not be large enough to represent the structural disorder in the absence of finite size effects. This ideal weak scaling efficiency breaks down when the problem size is instead increased in 3-D, which is a more relevant case for practical simulations. Here, new communication patterns are introduced across the $y$ and $z$ dimensions, and the connectivity extends to multiple neighbors, as can be inferred from the corresponding communication topology plots in \textbf{Fig. \ref{fig:weakscaling}(c)}. In \textbf{Fig. \ref{fig:weakscaling}(d)} we report the runtime decomposition for the data shown in \textbf{Fig. \ref{fig:weakscaling}(b)}. The time to pack data grows with increasing number of ranks, as more random accesses in the node embedding list are required. The NCCL send/recv time increases in turn from larger communication volume. These effects account for the loss of weak scaling efficiency in the 3-D case.

Overall, we achieve an 87\% weak scaling efficiency up to 128 nodes (512 GPUs) of Alps. In the case of 512 ranks, the average point-to-point communication volume per rank is 33.4 MiB (l$_{max}$ = 4, E = 16, $k$ $\sim$ 21k). Using the maximum NCCL send/recv time over all ranks, we measure an average bidirectional communication bandwidth of 20.4 GB/s, which corresponds to 42\% of the theoretical maximum \cite{hoefler_alps_bench}. The remaining unused bandwidth likely originates from variation in the number of embeddings sent to each neighboring rank - the volume of data which each rank must send to its neighbors decays sharply with the distance between their partitions of the graph. Communication time, meanwhile, is likely dominated by data exchanged between nearest neighbors.
 
\section{Modeling the electronic structure of GST}
\label{sec:insights}

As an example of the applications unlocked by performing ESP at large scales, we consider the case of phase change materials (PCM). These are compounds, typically GeSbTe or GST, which undergo transformations between crystalline and amorphous phases through (self-) heating \cite{burr_2010, LeGallo2020}. The heat to induce phase transitions can be delivered effectively though Joule heating under applied bias. In the presence of sufficient resistance contrast between phases, these materials can be used as memory devices. 

PCM have been previously commercialized as non-volatile memory \cite{Optane}, and are now being revisited for analog and in-memory computing \cite{KhaddamAljameh2021}. A full understanding of the coupling between atomic movements during phase change, electronic structure variation, and the resulting electrical current flow is key to overcoming limitations such as resistance drift \cite{LeGallo2020}. Having this insight can also enable computational explorations of the large stoichiometric design space of these compounds. The evolution of the atomic structure can in fact already be captured at scales of several tens of nanometers \cite{zhou2023}, matching the dimensions of fabricated devices. However, investigations of their electronic structure and transport properties have so far been limited to small structures where DFT-level computations are still feasible \cite{Holle2025, Chen2023}. 

\begin{figure}[t]
  \centering
  \includegraphics[width=\linewidth]{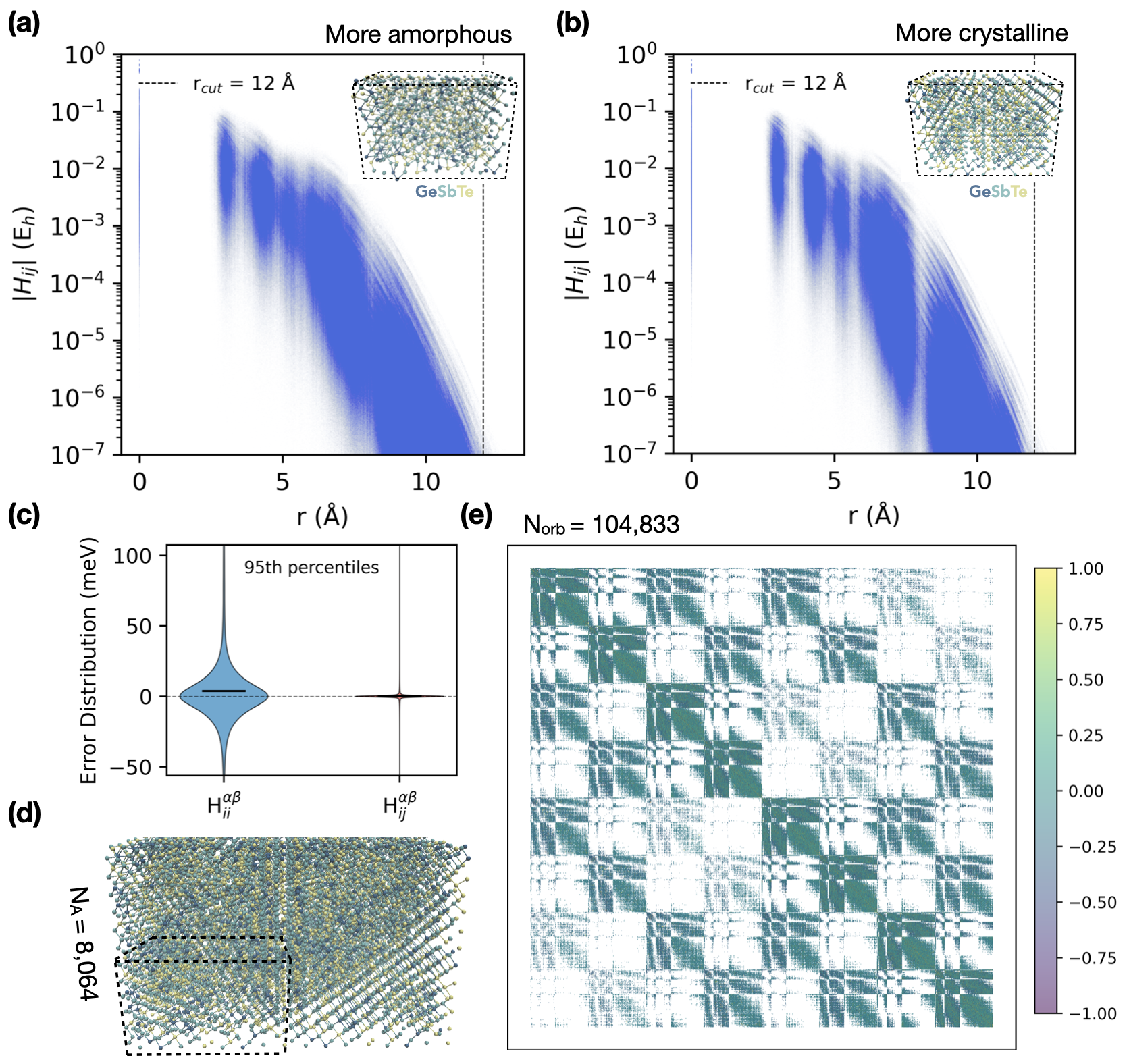}
  \caption{Learning the electronic structure of GeSbTe phase change materials. (a)/(b) Distribution of nonzero elements of $\mathbf{H}$ as a function of inter-atomic distance $r$, shown for (a) an amorphous and (b) a nearly-crystalline structure, each consisting of 1,008 atoms. Scatter points are plotted with a transparency of $\alpha$=0.01 to distinguish trends in the element magnitudes. The atomic structures corresponding to each case are pictured in the insets. (b) Distribution of mean absolute error of the structures used for validation (nearly crystalline). Black horizontal lines indicate the mean values of each violin. The y-axis limits are set to encompass all data from the 2.5th to the 97.5th percentile. The mean absolute error is shown separately for the nodes (diagonal blocks) and edges (off-diagonal blocks) as their elements differ in magnitude. (d) 2$\times$2$\times$2 tiled structure, consisting of 8,064 atoms, and (e) its predicted $\mathbf{H}$. Colors correspond to the log magnitude of each element.}
  \label{fig:GST}
\end{figure}

We therefore test our distributed eGNN model to learn the electronic structure of GST. We adapt atomic structures from the crystallization trajectory of a GST-124 system with 1,008 atoms, provided in \textbf{Ref.} \cite{zhou2023}, and then perform a structural relaxation and compute the $\mathbf{H}$ of selected snapshots. The Hamiltonian matrices of these structures were computed using a Double-$\zeta$ Valence Polarized (DZVP) basis with the CP2K DFT code \cite{cp2k}, which has 13 orbitals per atom for Ge, Sb, and Te (2$\times$s + 2$\times$p$_3$ + 1$\times$d$_5$). In \textbf{Fig. \ref{fig:GST}(a)-(b)} we show the magnitude of the elements of $\mathbf{H}$ for two snapshots extracted from the trajectory - one which is in the amorphous phase, and another closer to the crystalline phase. Crystallinity induces a narrow distribution of inter-atomic distances ($r$) between any two elements, leading to gaps in the available $\mathbf{H}$ elements as a function of $r$ (\textbf{Fig. \ref{fig:GST}(b)}). Meanwhile in the amorphous structure's $\mathbf{H}$, nearly every $r$ is sampled by the non-zeros in $\mathbf{H}$.

We use a combined L1+L2 loss over the elements of $\mathbf{H}$ \cite{quantumham} to train the eGNN network on the amorphous structure in \textbf{Fig. \ref{fig:GST}(a)}. r$_{cut}$ is set to 12.0 \AA, which is sufficient to capture almost all non-zeros of $\mathbf{H}$ (vertical dashed lines in \textbf{Fig. \ref{fig:GST}(a)-(b)}). The model contains three message passing layers. Full-batch distributed training was performed with a ReduceLRonPlateau learning rate scheduler, on 10 nodes (40 GPUs), for 7,700 optimization steps. The final errors for the diagonal and off-diagonal blocks of the `nearly crystalline' structure used for validation are shown in \textbf{Fig. \ref{fig:GST}(c)}. We find a mean absolute error of 2.64 meV, which is comparable in accuracy to similar ESP on materials with $<$150 atoms (2.2 meV \cite{universal_materials}). The structure was subsequently tiled 2$\times$ along all 3 dimensions, resulting in a larger system containing 8 replicas of the original unit cell (\textbf{Fig. \ref{fig:GST}(d)}). This final tiled cell contains 8,064 atoms and 1,832,304 edges. We use the trained network to construct its corresponding $\mathbf{H}$, which has a size of N=104,833. Although the structure is tiled, this approach still allows to test the network's ability to compute longer-range interactions than seen in the training data. The resulting matrix is shown in \textbf{Fig. \ref{fig:GST}(e)}. 




\section{Conclusion and Outlook}

We present a distributed GNN capable of processing the large, densely connected atomic graphs encountered in ESP tasks. By leveraging fast direct-GPU communication libraries and introducing an edge-wise `Low-NN' partitioning algorithm, we demonstrate weak scaling up to 512 GPUs on the Alps supercomputer. Our work introduces distributed training and inference of electronic ground-state Hamiltonian matrices for downstream device-level simulations at massive scales, enabling computational investigations of an extended class of electronic materials. Our implementation is made available at [repository shared upon publication]. Developing architectural and training improvements to further increase the accuracies of predicted matrices is now an active area of research.

Although the application and data processing pipeline presented here are tailored for electronic structure prediction, equivariant graph architectures in materials modeling were initially developed for force- and energy-prediction, and currently show state-of-the-art performance on these tasks \cite{equiformerv2, eSEN}. Our implementation can adapted for these properties, and applies broadly to this class of GNNs. Our methods may also be useful to the acceleration of several related ML applications in computational materials science. For example, the network can be trained to treat similarly tensor-valued properties of materials, such as dynamical matrices \cite{Okabe2024}, to serve as starting guesses for charge density during DFT \cite{wanet}, or to achieve more fine-grained parallelism on training datasets consisting of numerous smaller molecules.

\section*{Acknowledgements}
We acknowledge funding from the ALMOND project (SNSF Sinergia grant no. 198612) and from the Swiss State Secretariat for Education, Research, and Innovation (SERI) through the SwissChips research project. This work was supported by the Swiss National Supercomputing Center (CSCS) under projects lp16 and lp82.

\newpage
\section*{Appendices}
\appendix

\section{Low-NN Algorithm}

\begin{algorithm}[h]
\label{alg:domain_partitioning}
\nonl Let $N$ be the number of atoms in the domain

\SetKwFunction{cutdomain}{\textsc{Cut}-Domain}
\SetKwFunction{getcutpos}{\textsc{Get}-Cut}
\KwIn{$l$: partition level (2$^l$ total partitions)}
\KwIn{$r_{cut}$: cutoff radius}
\KwIn{$pos_{xyz}$ ($N \times 3$): atom positions (sorted)}
\KwIn{$N_D$ ($N \times 1$): degree (edges) of each atom}
\KwIn{$cuts$ ($3 \times 1$): \# cuts in each dimension}
\KwIn{$P_{o}$ ($k \times N_i \times 3$): list of $k$ partitions with $N_i$ atoms}
\KwData{$L$ ($3 \times 1$): (sub)domain size}
\KwData{$O$ ($3 \times 1$): (sub)domain origin} 
\KwData{$NN$: \# expected neighbours in each dimension} 
\KwData{$dim$: dimension to cut}

\vspace{5pt} 

\Indm\nonl\getcutpos{pos$_{xyz}$, N$_D$, $dim$, $O$}\\
\Indp
\vspace{5pt} 

Find $p$ such that $\sum_i$$N_D^{i<p}$ $\approx$ $\sum_i$$N_D^{i>p}$

Split $pos_{xyz}$ into:
\begin{itemize}
    \item Left partition: $pos_{xyz}^l$ ($pos_{xyz} < p$)
    \item Right partition: $pos_{xyz}^r$ ($pos_{xyz} > p$)
\end{itemize}


\Return{$p$, $pos_{xyz}^l$, $pos_{xyz}^r$}

\vspace{5pt} 

\Indm\nonl\cutdomain{$l, pos_{xyz}, N_D, P_o, r_{cut}, cuts$}\\
\Indp
\vspace{5pt} 

\If{$l=0$}{
    P$_o$.append(pos$_{xyz}$)\;
    \KwRet{}
}

\vspace{2pt} 

Compute L$_{xyz}$ from $pos_{xyz}$

\If{$\exists d \in \{0, 1, 2\} \text{ such that } cuts[d] = 0$}{
    $dim = \argmin\{cuts[d] \mid cuts[d] = 0\}$\;
    $O_{xyz} = \getcutpos(pos_{xyz}, N_D, dim, O)$\;
    $cuts[dim] \gets cuts[dim] + 1$\;
    \cutdomain{$l, pos_{xyz}, N_D, L, P_o, r_{cut}, cuts$}\;
}
\Else{
    $NN$ = [0,0,0]\;
    \For{$i \gets 0$ \KwTo $2$}{
        \Comment{Dimension is periodic}
        \If{$cuts[i] == 1$}{
            $NN$[i] = 1\;
        }
        \Comment{Dimension is not periodic}
        \Else{
            $NN$[i] = $\left\lceil \frac{2 \times r_{cut}}{L[i]} \right\rceil$\;
        }
    }

    \Comment{Cut dim with fewer neighbours}
    \If{$\text{argmin}(NN)$ \text{is unique}}{
        $dim = \text{argmin}(NN)$\;
    }
    \Else{
        $dim = \max(\text{argmin}(NN))$\;
    }

        $O$, $pos_{xyz}^{l}$, $pos_{xyz}^{r}$ = \getcutpos(pos$_{xyz}$, N$_D$, $dim$, $O$)\;

    $cuts[dim] \gets cuts[dim] + 1$\;
    \cutdomain{$l-1, pos_{xyz}^{l}, N_D, P_o, r_{cut}, cuts$}\;
    \cutdomain{$l-1, pos_{xyz}^{r}, N_D, P_o, r_{cut}, cuts$}\;
}

\end{algorithm}

\printbibliography

\end{document}